\documentclass[conference]{IEEEtran}
\IEEEoverridecommandlockouts
\usepackage{cite}
\usepackage{amsmath,amssymb,amsfonts}
\usepackage{algorithmic}
\usepackage{graphicx}
\usepackage{textcomp}
\usepackage{xcolor}

\def\BibTeX{{\rm B\kern-.05em{\sc i\kern-.025em b}\kern-.08em
    T\kern-.1667em\lower.7ex\hbox{E}\kern-.125emX}}

\usepackage{fancyhdr}
\begin{document}

\title{Student Classroom Behavior Detection and Analysis Based on Spatio-Temporal Network and Multi-Model Fusion 

}

\author{\IEEEauthorblockN{1\textsuperscript{st} FAN YANG}
\IEEEauthorblockA{\textit{Jinan University} \\
Guangzhou, China \\
winstonyf@qq.com}

\and
\IEEEauthorblockN{2\textsuperscript{nd} Xiaofei Wang}
\IEEEauthorblockA{\textit{Chengdu University} \\
Chengdu, China \\
wangxiaofei@cdu.edu.cn}

\and
\IEEEauthorblockN{3\textsuperscript{rd} Baocai Zhong}
\IEEEauthorblockA{\textit{Chengdu Neusoft University} \\
Chengdu, China \\
zhongbaocai@nsu.edu.cn}

\and
\IEEEauthorblockN{3\textsuperscript{th} Jialing	Zhong}
\IEEEauthorblockA{\textit{Chengdu Neusoft University} \\
Chengdu, China \\
zhongjialing@nsu.edu.cn}

}

\maketitle

\begin{abstract}
Using deep learning methods to detect students' classroom behavior automatically is a promising approach for analyzing their class performance and improving teaching effectiveness.   However, the lack of publicly available spatio-temporal datasets on student behavior, as well as the high cost of manually labeling such datasets, pose significant challenges for researchers in this field. To address this issue, we proposed a method for extending the spatio-temporal behavior dataset in Student Classroom Scenarios (SCB-ST-Dataset4) through image dataset.   Our SCB-ST-Dataset4 comprises 757265 images with 25810 labels, focusing on 3 behaviors: hand-raising, reading, writing. Our proposed method can rapidly generate spatio-temporal behavior datasets without requiring extra manual labeling. Furthermore, we proposed a Behavior Similarity Index (BSI) to explore the similarity of behaviors. We evaluated the dataset using the YOLOv5, YOLOv7, YOLOv8, and SlowFast algorithms, achieving a mean average precision (map) of up to 82.3$\%$. Last, we fused multiple models to generate student behavior-related data from various perspectives. The experiment further demonstrates the effectiveness of our method. And SCB-ST-Dataset4 provides a robust foundation for future research in student behavior detection, potentially contributing to advancements in this field. The SCB-ST-Dataset4 is available for download at: https://github.com/Whiffe/SCB-dataset. 
\end{abstract}


\section{Introduction}

\begin{figure}[htbp]
\centerline{\includegraphics[width=0.48\textwidth]{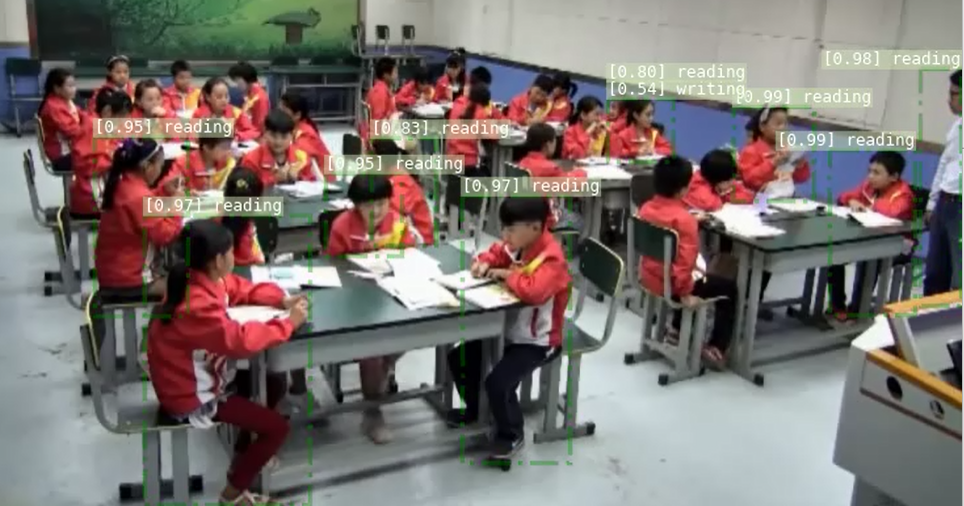}}
\caption{SlowFast detection in students' classroom.}
\label{SlowFast_demo}
\end{figure}

In recent years, behavior detection technology\cite{Zhu_comprehensive_study} has emerged as a valuable tool in analyzing student behavior in class videos (Fig.~\ref{SlowFast_demo}). This technology helps teachers, administrators, students, and parents understand classroom dynamics and learning performance. Traditional teaching models struggle to monitor every student's progress, relying on observations of a few students. Similarly, administrators and parents rely on limited information to assess education quality. Employing behavior detection technology to analyze student behavior accurately can offer more comprehensive and accurate feedback on education and teaching.

\begin{figure*}[htbp]
\centerline{\includegraphics[width=0.92\textwidth]{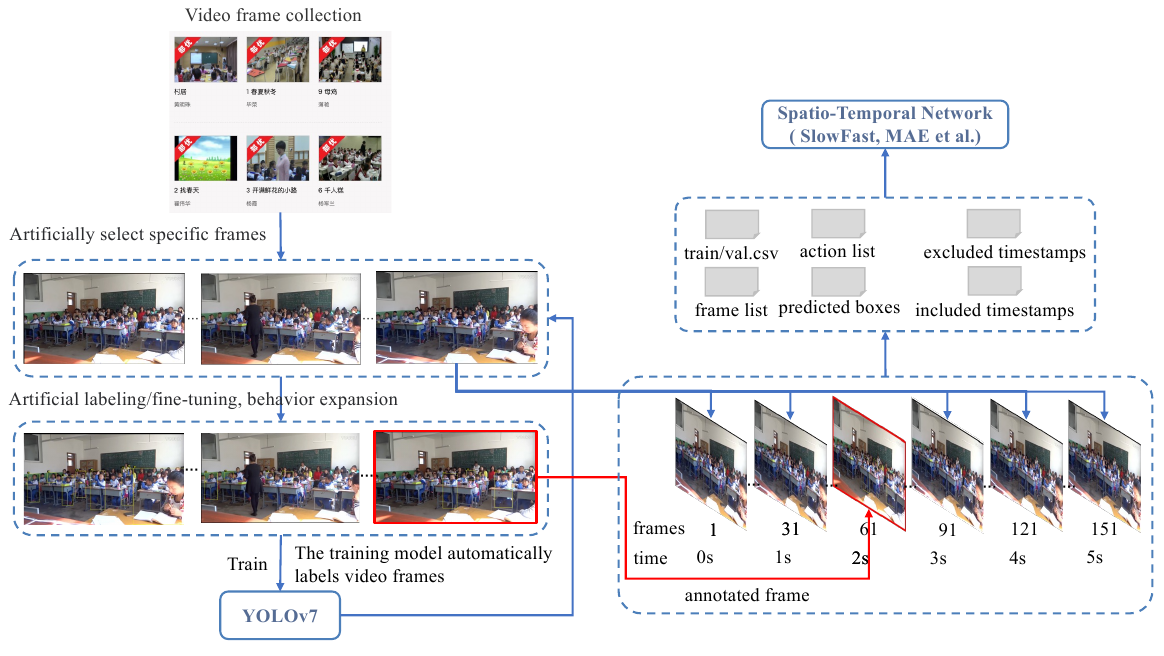}}
\caption{Process for creating SCB-ST-Dataset4.}
\label{Process_SCB-ST-Dataset4}
\end{figure*}
\textbf{Student classroom behavior datasets and methods}

Existing student classroom behavior detection algorithms can be roughly divided into three categories: video-action-recognition-based\cite{HUANG_Multi-person_classroom_action}, pose-estimation-based\cite{He_recognition_student_classroom_behavior} and object-detection-based\cite{YAN_Student_classroom_behavior}. Video-based student classroom behavior detection enables the recognition of continuous behavior, which requires labeling a large number of samples. For example, the AVA dataset\cite{Ava} for SlowFast\cite{Slowfast} and MAE\cite{MAE} detection is annotated with 1.58M. The number of annotations in datasets such as UCF101\cite{UCF101}, Kinetics400\cite{kinetics}, AVA-Kinetics\cite{AVA-kinetics}, and MultiSports\cite{Multisports} is also significantly large. Interestingly, in previous research\cite{Zhu_comprehensive_study} and experiments, similar findings have been observed:  some actions can sometimes be determined by the context or scene alone.  For example, the model can predict the action riding a bike as long as it recognizes a bike in the video frame.  The model may also predict the action cricket bowling if it recognizes the cricket pitch.  Hence for these classes, video action recognition may become an object/scene classification problem without the need of reasoning motion/temporal information. And pose-estimation-based algorithms characterize human behavior by obtaining the position and motion information of each joint in the body. Still, they are not applicable for behavior detection in overcrowded classrooms. in recent years object-detection-based algorithms have made tremendous breakthroughs, such as YOLOv5\cite{YOLOv5}, YOLOv7\cite{YOLOv7}, YOLOv8\cite{YOLOv8}. In the past, we attempted to employ the method of object-detection to achieve the detection of students' classroom behavior\cite{SCB-Dataset1, SCB-Dataset1-yolov7, SCB-Dataset2, SCB-Dataset3} and successfully achieved promising outcomes. 

In this study, we have iteratively optimized our previous work further to expand the SCB-Dataset\cite{SCB-Dataset1, SCB-Dataset1-yolov7, SCB-Dataset2, SCB-Dataset3}. Through comparing three behaviors (hand-raising, reading, writing), we aim to explore and analyze the strengths and weaknesses of object detection and spatio-temporal behavior detection in students' classroom behavior scenarios. This analysis can offer valuable insights for future researchers when selecting algorithms. 

In addition, we have conducted extensive data statistics and benchmark tests to ensure the dataset's quality, providing reliable Student Classroom Behavior Dataset. This work further addressed the research gap in detecting student behaviors in classroom teaching scenarios.

Our main contributions are as follows:

1. We proposed a method to expand the image dataset into a spatio-temporal behavior dataset(SCB-ST-Dataset4) using the SCB-Dataset3-U dataset. Our proposed method effectively reduces the manpower required and significantly increases the scale of video dataset creation.

2. We conducted extensive benchmark testing on the SCB-ST-Dataset4, providing a solid foundation for future research.

3. We proposed a new metric, the Behavior Similarity Index (BSI), which measures the similarity in form between different behaviors under a network model.

4.We fused multiple models, including Deep Sort, YOLOv7, SynergyNet, and Facial Expression. The data from these multiple models is important for assisting in the analysis of student behavior

4.We propose a student concentrated behavior sequence analysis process that can analyze the duration of concentrated and unconcentrated for each student throughout the video, and calculate their concentrated scores.

\section{Related Works}

Recently, many researchers have utilized computer vision to detect student classroom behaviors. However, the lack of public student behavior datasets in education restricts the research and application of behavior detection in classroom scenes. Many researchers have also proposed many unpublished datasets. The ActRec-Classroom dataset \cite{Fu_behavior_analysis} includes 5 classes (listening, fatigue, raising hand, leaning, and reading/writing) with 5,126 images, and the action recognition method involves using Faster R-CNN for detecting human bodies, OpenPose for extracting key points, and a CNN-based classifier for classification. A large-scale dataset for student behavior \cite{Zheng_student_behavior}, collected from thirty schools and labeled using bounding boxes frame-by-frame, includes 70k hand-raising samples, 20k standing samples, and 3k sleeping samples. An enhanced Faster R-CNN model for student behavior analysis incorporates a scale-aware detection head, a feature fusion strategy, and the use of OHEM to improve detection performance while reducing computation overhead and addressing class imbalances in a real corpus dataset. BNU-LCSAD \cite{Sun_Student_Class_Behavior_Dataset} is a comprehensive dataset with 128 videos from 11 classrooms in different disciplines, which can be used for recognizing, detecting, and captioning students' behaviors, with baseline models including the two-stream network, ACAM \cite{ACAM}, MOC \cite{MOC}, BSN \cite{Bsn}, DBG \cite{DBG}, RecNet \cite{RecNet}, and HACA \cite{HACA}. The Student Classroom Behavior Dataset \cite{Zhou_Classroom_Learning} consists of 90 classroom videos capturing 400 students in a primary school. It includes 10,000 images of students engaging in activities such as raising hands, walking, writing on the blackboard, and looking up and down, as well as 1,000 images of students bending down, standing, and lying on the table. The dataset is utilized with a 10-layer deep convolutional neural network (CNN-10) to recognize student classroom behaviors by extracting key information from human skeleton data, effectively excluding irrelevant information, and achieving higher recognition accuracy and generalization ability. A dataset of student behavior \cite{Li_Student_behavior} in an intelligent classroom was created, which includes seven classes of behavior and utilizes challenging class surveillance videos for annotation. The method integrates relational features to analyze interactions between actors and their context, modeling human-to-human relationships using body parts and context for accurate interaction recognition. A dataset \cite{Trabelsi_Classroom} of 3881 labeled images capturing diverse student behaviors in a classroom setting, such as raising hands, paying attention, eating, being distracted, reading books, using phone, writing, feeling bored, and laughing, is utilized for training and evaluation using the YOLOv5 object detection model. The large-scale student behavior dataset \cite{Zhou_Stuart} contains five representative student behaviors highly correlated with student engagement and tracks the change trends of these behaviors during the course, and it is utilized in the proposed StuArt, an innovative automated system that enables instructors to closely monitor the learning progress of each student in the classroom, including user-friendly visualizations to facilitate instructors' understanding of individual and overall learning status. The classroom behavior dataset \cite{BiTNet} consists of 4432 images with 151574 annotations, capturing 7 common student behaviors and 1 typical teacher behavior, and it is utilized in the proposed BiTNet, a real-time object detection network aimed at enhancing teaching quality and providing real-time analysis of student behavior in the classroom, addressing challenges such as occlusion and small objects in images. The Teacher Behavior Dataset (STBD-08) \cite{CBPH-Net} contains 4432 images with 151574 labeled anchors covering eight typical classroom behaviors, and to address challenges such as occlusions, pose variations, and inconsistent target scales, the authors propose an advanced single-stage object detector called ConvNeXt Block Prediction Head Network.

\section{SCB-ST-Dataset4}

\subsection{The SCB-ST-Dataset4 Production Process}

Understanding student behavior is crucial for comprehending their learning process, personality, and psychological traits and is important in evaluating the quality of education. The hand-raising, reading, writing are important indicators of evaluating classroom quality.    Based on our previous work, Based on the SCB-Dataset3-U dataset, we used the method shown in the Fig.~\ref{Process_SCB-ST-Dataset4} to expand the image dataset into a spatio-temporal behavior dataset.  This expansion method effectively reduces the required manpower and significantly increases the scale of video dataset creation. This expanded dataset is referred to as SCB-ST-Dataset4.

 For the SCB-ST-Dataset4 (classroom dataset from kindergarten to high school), we collected over a thousand videos ranging from kindergarten to high school, each lasting approximately 40 minutes. We intentionally selected 3 to 15 frames with a specific time interval for each video to reduce imbalances in behavior classes and improve representativeness and authenticity. The videos were extracted from the ``bjyhjy'', ``1s1k'', ``youke''.  ``qlteacher'' and ``youke-smile.shec'' websites.

The production of SCB-ST-Dataset4 is divided into two parts.  The first part entails the content on the left side of Fig.~\ref{Process_SCB-ST-Dataset4} , which pertains to the production process of the image dataset.  The right side of the figure outlines the procedure for expanding the image dataset into a spatio-temporal behavior dataset.  

For the left part, firstly, we collected a series of suitable video frames from an online course website and carefully screened them.  Next, we manually annotated these video frames to provide accurate label data for subsequent training.  Then, we input the annotated data into the YOLOv7 model for training.  By using this model, we can perform inference detection on the unannotated video frames, the model detects the target objects in the videos.  We further fine-tune these detection results using artificial intelligence techniques to improve the accuracy and reliability of the detection while reducing human effort.  Finally, we input the fine-tuned data back into the YOLOv7 model for training again to further optimize the model's performance.  Through iterative training, we continuously improve the accuracy and robustness of the model, resulting in more accurate detection results.

For the right part, each video in our extended spatio-temporal behavior dataset is set to a length of 5 seconds (30 frames per second). First, we select a video frame with existing labels as the frame for the 2nd second (61st frame). Then, a crucial step is taken: if we are able to track the position of this video frame in the video, we extract the preceding 60 frames and the subsequent 90 frames to form a continuous spatio-temporal behavior video. If we are unable to track the position of the video frame in the video, we replicate the video and make 60 copies in front and 90 copies at the back. Next, we generate labels for the spatio-temporal behavior dataset and input them into the spatio-temporal behavior detection network for training.

\begin{figure}[htbp]
\centerline{\includegraphics[width=0.45\textwidth]{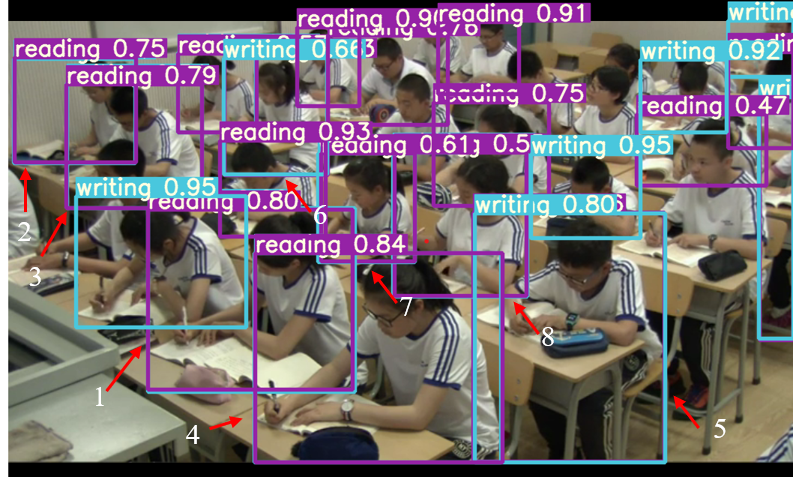}}
\caption{ Overlapping results of detection for reading and writing behavior.}
\label{class_overlap}
\end{figure}

\subsection{The SCB-ST-Dataset4 Challenge}

During the iterative training process on SCB-ST-Dataset4, we encountered an issue where there was an overlap between the bounding boxes of the ``reading'' and ``writing'' behaviors in the behavior detection (see Fig.~\ref{class_overlap}). Specifically, there were 8 instances where the bounding boxes overlapped between these two behaviors. This observation indirectly suggests that ``reading'' and ``writing'' behaviors are visually similar, making distinguishing between them more challenging for the YOLOv7 model. To improve the accuracy of the model, further exploration is needed to address this challenge.

\begin{figure}
\centerline{\includegraphics[width=0.48\textwidth]{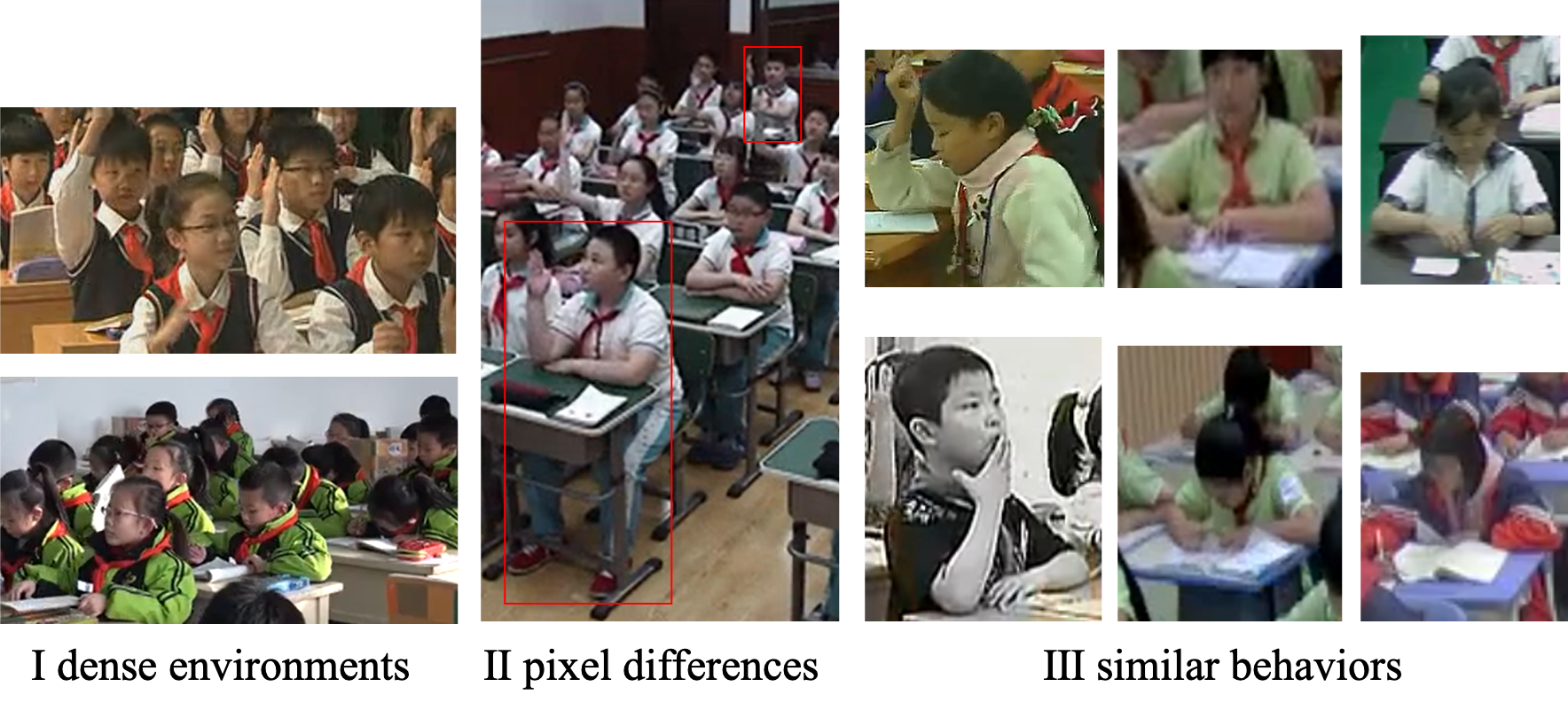}}
\caption{Challenges in the SCB-ST-Dataset4 include dense environments, similar behaviors, and pixel differences..}
\label{dense_pixel_similar}
\end{figure}

And the classroom environment presents challenges for detecting behavior due to the crowded students and variation in positions, as shown in Fig.~\ref{dense_pixel_similar} \uppercase\expandafter{\romannumeral1} and \uppercase\expandafter{\romannumeral2}. In addition to the high degree of similarity between reading and writing in certain scenarios, there is also a visual similarity between hand-raising and other classes of behavior. This further complicates the detection process, as shown in Fig.~\ref{dense_pixel_similar} \uppercase\expandafter{\romannumeral3}.

\begin{figure}
\centerline{\includegraphics[width=0.48\textwidth]{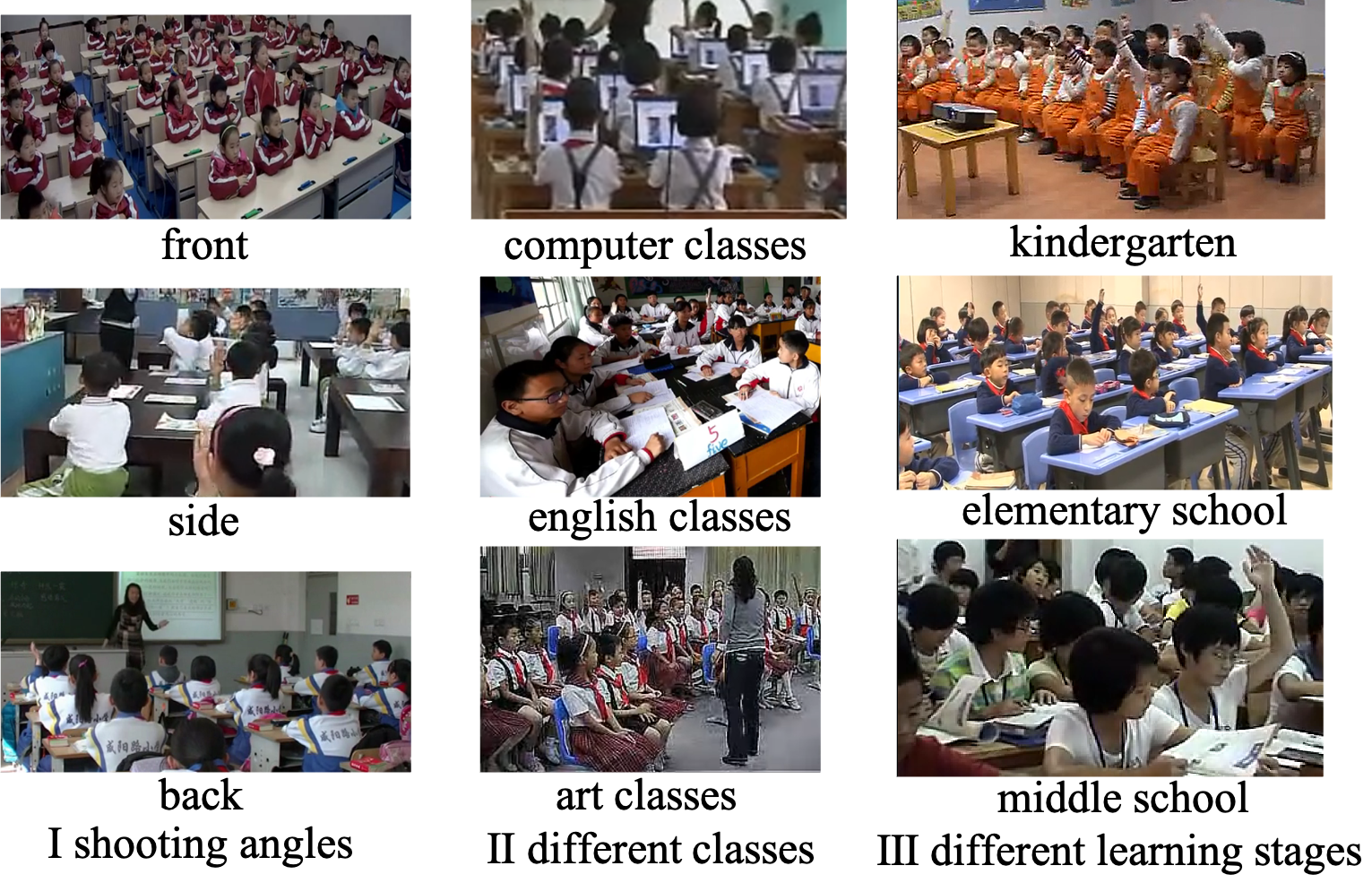}}
\caption{Challenges in the SCB-ST-Dataset4 include varying shooting angles, class differences, and different learning stages.}
\label{angles_class_stages}
\end{figure}

The SCB-ST-Dataset4 was collected from different angles, including front, side, and back views (Fig.~\ref{angles_class_stages} \uppercase\expandafter{\romannumeral1}). Additionally, the classroom environment and seating arrangement can vary, which adds complexity to the detection task (Fig.~\ref{angles_class_stages} \uppercase\expandafter{\romannumeral2}). Student classroom behavior behaviors also differ across learning stages from kindergarten to high-school, creating challenges for detection (Fig.~\ref{angles_class_stages} \uppercase\expandafter{\romannumeral3}).

\subsection{Statistics of The SCB-ST-Dataset4}

\begin{table}
    \small
    \centering \renewcommand\arraystretch{1.25} 
    \caption{comparisons of SCB-Dataset1, SCB-Dataset2, SCB-Dataset3-S, SCB-Dataset3-U, SCB-ST-Dataset4.}
    \label{comparisons_SCB123}
    \begin{tabular}{p{1.2cm}p{0.9cm}p{0.9cm}p{1.1cm}p{1.1cm}p{1.3cm}}
        \hline
         & SCB1 & SCB2 & SCB3-S & SCB3-U & SCB4 \\ \hline
        classes & 1 & 3  & 3 & 6 & 3 \\
        Images & 4001 & 4266 & 5015 & 671 & 757265\\
        Annotations & 11248 & 18499 & 25810 & 19768 & 25810 \\
        \hline
    \end{tabular}
\end{table}

Table  \ref{comparisons_SCB123} shows the one-year iteration process of our student behavior dataset, where SCB1 represents SCB-Dataset1, SCB2 represents SCB-Dataset2, SCB3-S represents SCB-Dataset3-S, SCB3-U represents SCB-Dataset3-U, and SCB4 represents  SCB-ST-Dataset4 . The ``class'' row indicates the behaviors. The number 1 represents the hand-raising behavior, the number 3 represents the hand-raising, reading, and writing behaviors, and the number 6 represents the hand-raising, reading, writing, using phone, bowing the head, and leaning over the table. It can be seen from the table that our dataset has increased in both class and image quantity, especially notable is the rapid expansion of SCB-ST-Dataset4 in terms of image quantity.

\begin{table}
    \small
    \centering \renewcommand\arraystretch{1.25} 
    \caption{Statistics of behavior classes in different datasets.}
    \label{Statistics_behavior_classes}
    \begin{tabular}{ccccccc}
        \hline
         & hand-raising & reading & writing  \\ \hline
        SCB1 & 11248 & - & -  \\
        SCB2 & 10078 & 5882 & 2539  \\
        SCB3-S & 11207 & 10841 & 3762 \\
        SCB4 & 11207 & 10841 & 3762  \\
        \hline
    \end{tabular}
\end{table}

Table \ref{Statistics_behavior_classes} displays the number of different behavior classes in each dataset.

\begin{figure}[htbp]
\centerline{\includegraphics[width=0.48\textwidth]{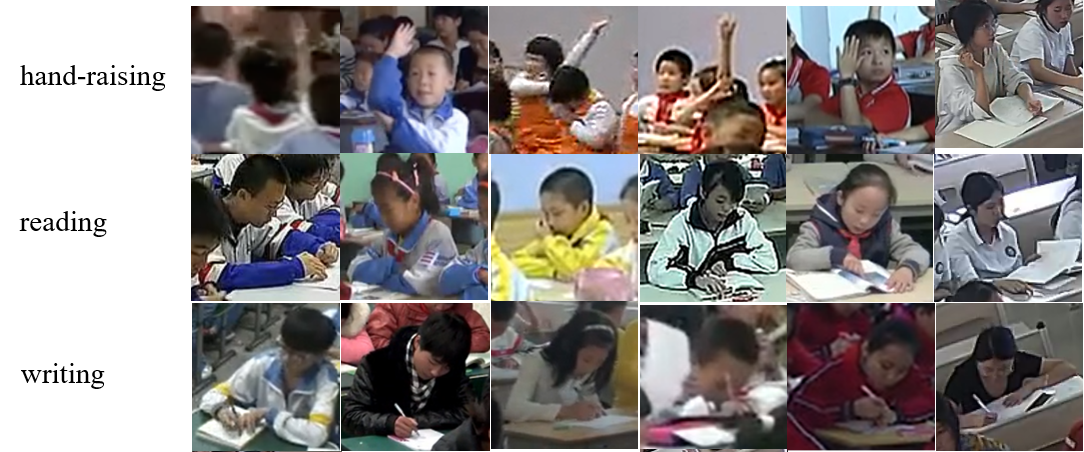}}
\caption{Example images of classroom behavior dataset.}
\label{Example_images_of_classroom_behavior_dataset}
\end{figure}

Some visual examples of the SCB-ST-Dataset4 are illustrated in Fig.~\ref{Example_images_of_classroom_behavior_dataset} 

\section{Spatio-Temporal Network and Multi-Model Fusion}

\subsection{Spatio-Temporal Network}

\begin{figure}[htbp]
\centerline{\includegraphics[width=0.45\textwidth]{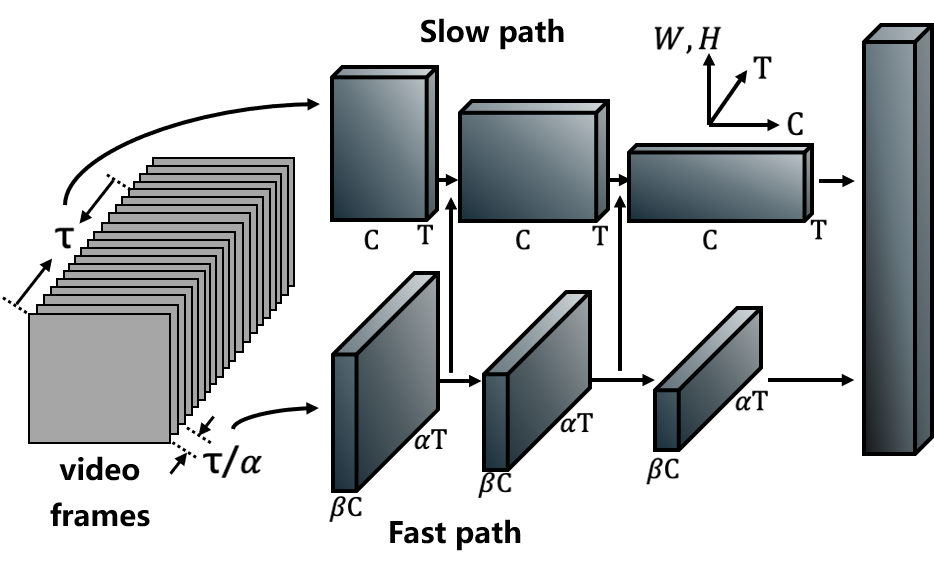}}
\caption{The network structure of the SlowFast.}
\label{SlowFastNet}
\end{figure}

SlowFast\cite{Slowfast} is a dual-stream network structure. One stream processes the semantic information of the video's spatial features at a low frame rate (Slow path, with frames input every $\tau$ frames, where $\tau =16$). The other stream processes the motion information at a high frequency (Fast path, with frames input every $\tau/\alpha$ frames, where $\alpha=8$). As shown on the left side of Fig. \ref{SlowFastNet}, the feature channels for the Slow path are represented by C, while the Fast path has $\beta C$ feature channels, where $\beta=1/8$, allowing the Fast path to run at a higher speed. In terms of time dimensions, the Slow path has a time dimension of T, while the Fast path has $\alpha$ T to better capture motion information. The paths share the same H and W dimensions to allow for lateral connections. SlowFast merges the features of the dual-stream branches via multiple lateral connections, and the merged information is then fed into a classifier for multi-label classification prediction.

\subsection{Multi-Model Fusion}

\begin{figure*}[htbp]
\centerline{\includegraphics[width=0.92\textwidth]{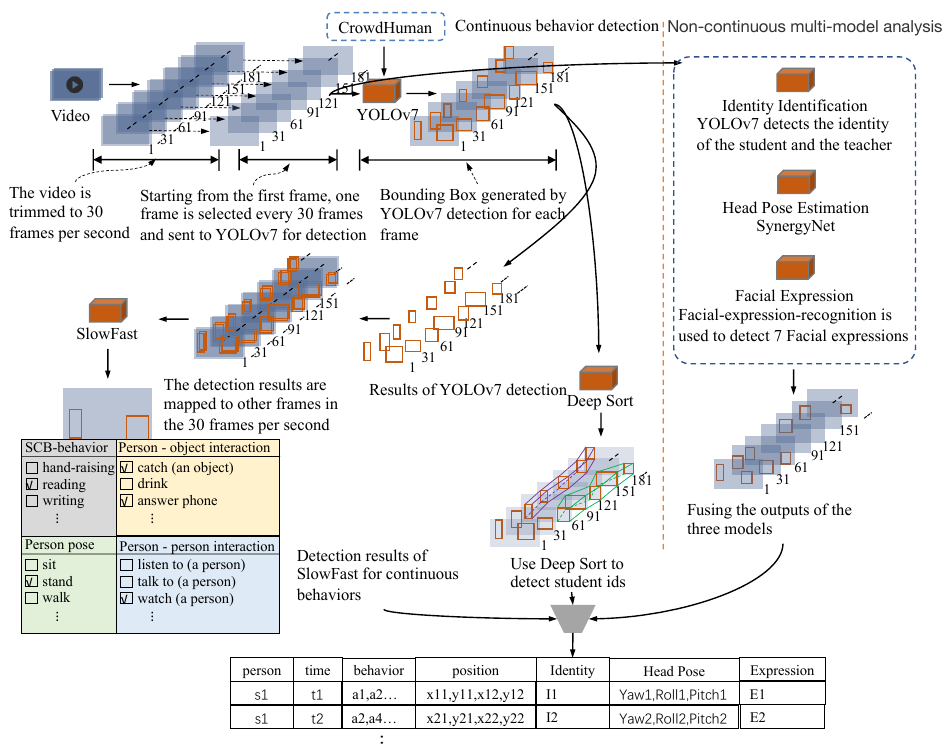}}
\caption{Student Behavior Analysis System with Multi-model Fusion.}
\label{Multi_Model_Fusion}
\end{figure*}

Our multi-model fusion system for student behavior analysis is shown in Fig.~\ref{Multi_Model_Fusion}. It consists of three parts: continuous behavior detection, non-continuous multi-model analysis, and the fusion of multi-model results.

To detect continuous student behavior, we sampled the video at 30 frames per second and used YOLOv7 with weights trained on the CrowdHuman dataset~\cite{Crowdhuman} to detect Person every 30 frames. The detection results were sent to Deep Sort and Spatio-Temporal Network(SlowFast). SlowFast detected continuous behaviors, We classified continuous behaviors into four major categories: student classroom behaviors (such as hand-raising, reading, writing, etc.), person pose (such as sit, stand, walk etc.), and person-person interaction (listen, talk, watch etc.). The student classroom behaviors are customized by us, and the remaining three categories come with SlowFast. Deep Sort is used to determine the ID of the person in the video.

For non-continuous multi-model analysis, the video was sampled at 1 frame per second. We used three models: YOLOv7 was employed to detect the identity of the student and the teacher, SynergyNet~\cite{SynergyNet} was used for head pose estimation, and the Facial Expression model was implemented to recognize seven expressions: anger, disgust, fear, happiness, neutrality, sadness, and surprise.

Finally, we fuse the results of spatio-temporal behavior detection, Deep Sort, YOLOv7, SynergyNet, and the Facial Expression model to form a data table. The data table contains a row for each student/teacher, storing their multimodal result data.

We utilized YOLOv7 for student detection, but found it performed poorly in crowded classroom settings. Therefore, we adapted YOLOv7 to these settings by using weights trained on the CrowdHuman dataset.

\subsection{Student Concentration Analysis}

\begin{figure*}[htbp]
\centerline{\includegraphics[width=0.92\textwidth]{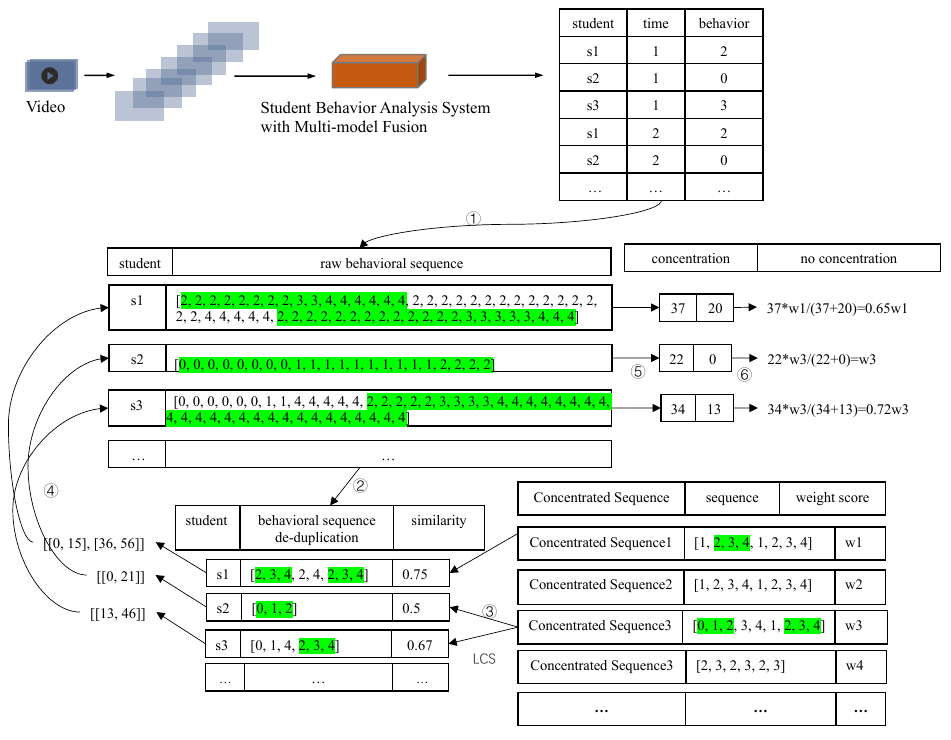}}
\caption{Student Behavioral Concentration Analysis Process.}
\label{concentration_analysis}
\end{figure*}

Fig.~\ref{concentration_analysis} illustrates the process of analyzing student behavior concentration. We start with the data outputted by Student Behavior Analysis System with Multi-model Fusion, including student ID, time ID, and behavior ID. We then reorganize these behavioral data, as shown in step 1 of Fig.~\ref{concentration_analysis}, by grouping the behaviors of each student into raw behavioral sequences throughout the entire video. It should be noted that these raw behavioral sequences often contain a lot of repetition due to the continuity of behaviors. Next, we perform duplicate removal on the raw behavioral sequences, as shown in step 2 in Fig. 9, to obtain a behavioral sequence without duplicates.

In the bottom right corner of Fig.~\ref{concentration_analysis}, we present a set of concentrated sequences. These sequences are determined based on specific scenarios such as student self-study class, student interactive class, student experimental class, and student math class, where each scenario corresponds to a different concentrated sequence. These concentrated sequences can be proposed by experts in the field of education, and each sequence is assigned a corresponding weight score, w.

Next, in step 3 of Fig.~\ref{concentration_analysis}, we use the Longest Common Subsequence (LCS) algorithm to find the most similar sequence between the student behavioral sequences and the concentrated sequence group (It is also possible that a matching sequence may not be found), resulting in a similarity value. For example, in Fig.~\ref{concentration_analysis}, the similarity value between student s1 and concentrated sequence 1 is 0.75. Then, through traversal, we find the sequence values that have an intersection between the student behavioral sequences and the concentrated sequences (the matching length needs to be greater than or equal to 2), as shown in Fig.~\ref{concentration_analysis}, the intersected sequence of student s1 and concentrated sequence 1 is [2, 3, 4]. We have marked these intersected sequences with a green background.

Next is step 4 of Fig.~\ref{concentration_analysis}, where we map the intersected sequences back to the raw behavioral sequences. We employ the method of traversing through a subset sliding window to find the starting and ending points of the intersected sequences in the original sequences, achieving data mapping. In Fig.~\ref{concentration_analysis}, we have represented the mapped values with a green background.

Finally, in step 5 of Fig.~\ref{concentration_analysis}, we calculate the length of the concentrated sequences and the length of the unconcentrated sequences. Then, in step 6 of Fig.~\ref{concentration_analysis}, we compute the concentration score of the student behavioral sequences. The calculation formula is: concentrated * w / (concentrated + unconcentrated).

\section{Experiments}

\subsection{Experimental Environment and Dataset}

The experiment was compiled and tested using Python 3.8, the corresponding development tool was PyCharm, the main computer vision library was python-OpenCV 4.1.2, the deep learning framework used was Pytorch v1.11 with CUDA version 11.3 for model training, the operating system was Ubuntu 20.04.2, and the CPU: 12 vCPU Intel(R) Xeon(R) Platinum 8255C CPU @ 2.50GHz.  The GPU: RTX 3080(10GB).

The dataset used in our experiments is SCB-Dataset3, SCB-ST-Dataset4 and StudentTeacher-Dataset, which we split into training, validation sets with a ratio of 4:1.

\subsection{Model Training}

For SCB-Dataset3, the model was trained with an epoch set to 100, batch size set to 8, and image size set to 640x640. A pre-trained model was utilized for the training process.

As for SCB-ST-Dataset4, the model was trained with an epoch set to 43, batch size set to 2, learning rate set to 0.1, and image size set to 640x640. Similarly, a pre-trained model was used for the training.

\subsection{Evaluation Metric}

When evaluating the results of an experiment, we use two main criteria: Precision and Recall. True Positive (TP) represents a correct identification, False Positive (FP) means an incorrect identification, and False Negative (FN) indicates that the target was missed.

To calculate Precision (Eq. \ref{precision}), we divide the number of True Positives by the sum of True Positives and False Positives. For Recall (Eq. \ref{recall}), we divide the number of True Positives by the sum of True Positives and False Negatives. Both Precision and Recall must be considered to properly assess the accuracy of the experiment.

\vspace{-2ex}
\begin{equation}
    \small
    \centering
    \begin{aligned}
        precision~ = ~\frac{TP}{TP + FP}
    \end{aligned}
    \label{precision}
\end{equation}

\vspace{-2ex}
\begin{equation}
    \small
    \centering
    \begin{aligned}
        recall~ = ~\frac{TP}{TP + FN}
    \end{aligned}
    \label{recall}
\end{equation}
\vspace{-2ex}

 To provide a more thorough evaluation of Precision and Recall, the metrics of Average Precision (AP, Eq. \ref{AP}) and mean Average Precision (mAP, Eq. \ref{mAP}) have been introduced.  These metrics calculate the average Precision over a range of Recall values, which provides a more comprehensive assessment of the model's performance.  AP is the average of Precision values at all Recall levels, and mAP is the mean AP value averaged over different classes.

\vspace{-2ex}

 \begin{equation}
    \small
    \centering
    \begin{aligned}
        AP_{i} = {\int_{0}^{1}{P(r)dr}}\\
    \end{aligned}
    \label{AP}
\end{equation} 
\vspace{-5ex}

\begin{equation}
    \small
    \centering
    \begin{aligned}
        mAP = \frac{1}{n}{\sum\limits_{i}^{n}\left( AP_{i} \right)}
    \end{aligned}
    \label{mAP}
\end{equation}
\vspace{-2ex}

As shown in Table \ref{comparison_of_YOLO_on_SCB-Dataset3-S}, mAP@50 refers to the mean average precision at IoU threshold 0.5, while mAP@50:95 represents the mean average precision at IoU thresholds ranging from 0.5 to 0.95.

\vspace{-2ex}
\begin{equation}
    \small
    \centering
    \begin{aligned}
    BSI = ( \frac{S_{i,j}}{N_{i}},  \frac{S_{i,j}}{N_{j}})
    \end{aligned}
    \label{BSI}
\end{equation}
\vspace{-2ex}

To calculate the similarity of student behaviors, we introduce the BSI (Behavior Similarity Index), as shown in Eq. \ref{BSI}. In the equation, $S_{ij}$ represents the overlap count between behavior class i and behavior class j,  $N_{i}$ represents the count of behavior class i, and $N_{j}$ represents the count of behavior class j. Therefore, BSI represents the overlap rate of behavior classes i and j, and BSI can to some extent reflect the similarity between behaviors.

 \begin{table}
    \small
    \centering \renewcommand\arraystretch{1.25} 
    \caption{The comparison of YOLOv5, YOLOv7, YOLOv8 on SCB-Dataset3-S.}
    \label{comparison_of_YOLO_on_SCB-Dataset3-S}
    \begin{tabular}{p{1.9cm}p{0.35cm}p{0.35cm}p{0.35cm}p{0.35cm}p{0.35cm}p{0.35cm}p{0.35cm}}
        \hline
         Models & v5s & v5m & v7 & v7x & v8s & v8m & SF  \\ \hline
        mAP@ 50(\%) & 72.9 & 74.7 & 77.2 & 80.3 & 74.5 & 77.6 & \underline{\textbf{82.3}}  \\
       
        \hline
    \end{tabular}
\end{table}

\subsection{Analysis of spatio-temporal detect experimental results}

For our training on SCB-Dataset3 and SCB-ST-Dataset4, we employed YOLOv5, YOLOv7 YOLOv8, SlowFast, The outcomes of our experiments are detailed in Table \ref{comparison_of_YOLO_on_SCB-Dataset3-S}. In the table,, v5s represents YOLOv5s, v5m represents YOLOv5m, v7 represents YOLOv7, v7x represents YOLOv7x, v8s represents YOLOv8s, v8m represents YOLOv8m, and SF represents SlowFast. From the table, it can be seen that SlowFast achieves the highest mAP@50, which is 82.3$\%$.

\begin{figure}[htbp]
\centerline{\includegraphics[width=0.48\textwidth]{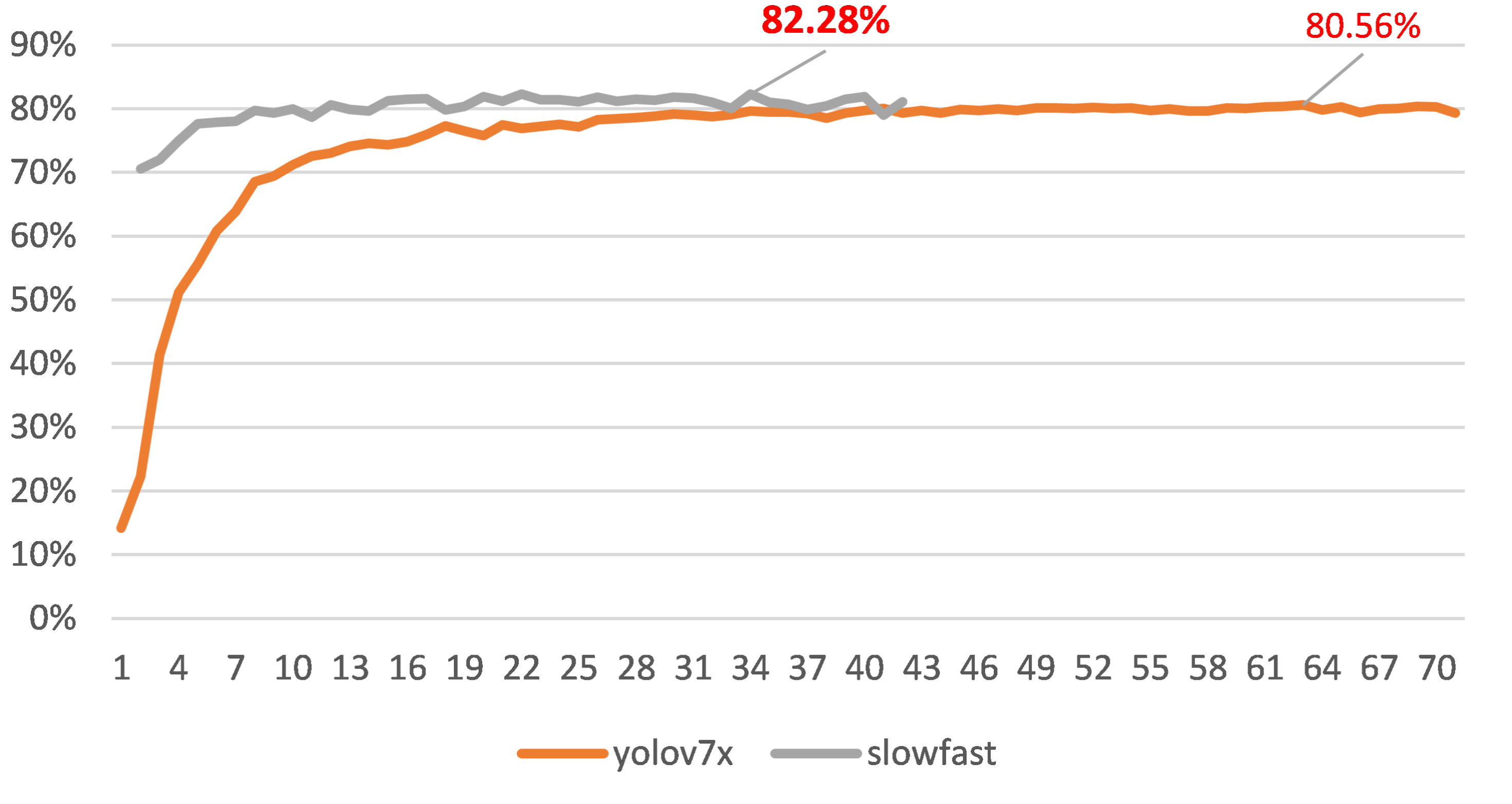}}
\caption{Comparison of mAP@50 during training of YOLOv7 and SlowFast.}
\label{train_yolo_slowfast}
\end{figure}

We compared the version of YOLO with the highest training accuracy in the YOLO series to SlowFast. As shown in Fig.~\ref{train_yolo_slowfast}, SlowFast iterated a total of 43 times, while YOLOv7x iterated 100 times. From the graph, it can be seen that SlowFast quickly converges and achieves an accuracy of over 80$\%$ within the first 10 iterations. Whether it is from the convergence speed or the final accuracy, SlowFast outperforms YOLOv7.

\begin{table}
    \small
    \centering \renewcommand\arraystretch{1.25} 
    \caption{Detection Results(AP@50) of Various classes in SCB-ST-Dataset4 using YOLOv5m, YOLOv7x, YOLOv8m, and SlowFast}
    \label{Various_classes_on_YOLO_SlowFast}
    \begin{tabular}{p{1.7cm}p{2.2cm}p{1.6cm}p{1.6cm}}
        \hline
         Models & hand-raising (\%)   &  reading (\%) &  writing (\%)   \\  
           \hline
        YOLOv5m & 85.4 & 75.2 & 63.3 \\
        YOLOv7x & 86.8 & 77.4 & \underline{\textbf{70.0}} \\
        YOLOv8m & 84.6 & 76.2 & 66.8 \\
        SlowFast & \underline{\textbf{96.9}} & \underline{\textbf{89.8}} & 57.8  \\
        \hline
    \end{tabular}
\end{table}

In order to further analyze the detection accuracy of different networks for each behavior (with data imbalance between behaviors), using the weights from the last training weight, we compared the accuracy of YOLOv5m, YOLOv7x, YOLOv8m, and SlowFast models on three behaviors as shown in Table \ref{Various_classes_on_YOLO_SlowFast}. It can be seen that compared to SlowFast, the YOLO series has better detection performance for the behavior "writing" with less data. This is because during the training process, the YOLO series weights the imbalanced data, allowing the network to learn more from the minority samples, while SlowFast does not perform this operation.

\begin{figure}[htbp]
\centerline{\includegraphics[width=0.48\textwidth]{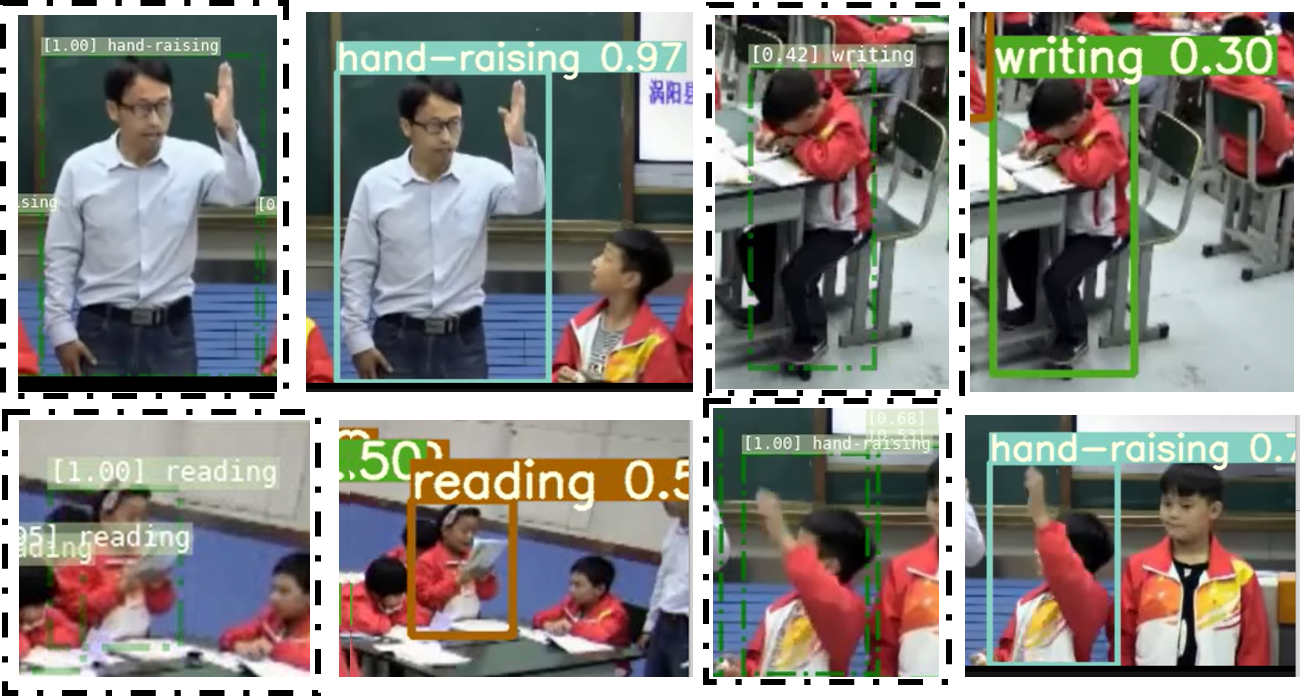}}
\caption{Comparison of SlowFast and YOLOv7x Detection.}
\label{YOLOv7x_SlowFast}
\end{figure}

 Fig.~\ref{YOLOv7x_SlowFast} demonstrates a comparison of the detection results between SlowFast and YOLOv7x on the same video. The results generated by SlowFast are enclosed within dashed lines.

\begin{table}
    \small
    \centering \renewcommand\arraystretch{1.25} 
    \caption{Results of BSI for different network models}
    \label{different_network_BSI}
    \begin{tabular}{ccccccccc}
        \hline
         Models& BSI(0,1) & BSI(0,2) &  BSI(1,2)   \\ \hline
        YOLOv5n & ( 0.5, 0.4 )  & ( 0.1, 0.2 ) & ( 3.8, 14.1 )  \\
        YOLOv5s & ( 0.5, 0.4 ) & ( 0.2, 0.5 ) & ( 3.3, 11.9 )  \\
        YOLOv5m & ( 0.4, 0.4 ) & ( 0.3, 0.8 ) & ( 3.2, 11.8 )  \\
        YOLOv7 &  ( 0.7, 0.6 ) &  ( 0.1, 0.3 ) & ( 5.7, 18.6 )  \\
        YOLOv7x & ( 0.7, 0.5 )  & ( 0.0, 0.1 ) & ( 4.9, 17.2 ) \\
        YOLOv8n & ( 0.3, 0.3 ) & ( 0.0, 0.1 ) & ( 4.5, 15.8 ) \\
        YOLOv8s & ( 0.3, 0.3 ) &  ( 0.0, 0.1 ) & ( 4.2, 15.5 ) \\
        YOLOv8m & ( 0.4, 0.3 ) & ( 0.1, 0.2 ) & ( 3.6, 12.9 )   \\
        \hline
    \end{tabular}
\end{table}

In addition, we conducted experiments to validate the effectiveness of the BSI proposed by us. Through visual observation, we found that writing and reading are similar behaviors. Therefore, we proposed the BSI to quantify this similarity and recorded the experimental results in Table \ref{different_network_BSI}, we used 0 to represent hand-raising, 1 to represent reading, and 2 to represent writing. Based on the results from Table \ref{different_network_BSI}, we can observe that hand-raising behavior has almost no similarity with other behaviors while reading and writing exhibit a higher degree of similarity.

\subsection{Analysis of multi-model experimental results}

\begin{figure}[htbp]
\centerline{\includegraphics[width=0.48\textwidth]{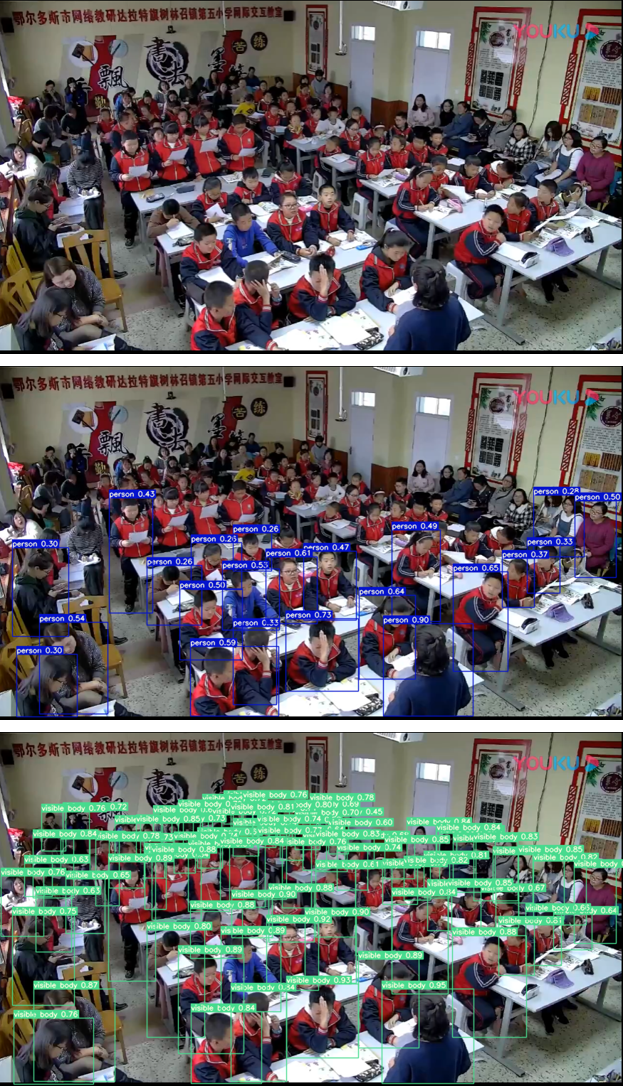}}
\caption{Multi model detection results}
\label{yolov7_crowdhuman_compare}
\end{figure}

\textbf{YOLOv7+CrowdHuman}
We utilized the YOLOv7x network (with a pre-trained model) and trained it for 200 epochs on the CrowdHuman dataset. The mAP@50 reached 85.6$\%$(Table \ref{yolov7_crowdhuman}). The comparison of model detection performance before and after training is shown in Fig.~\ref{yolov7_crowdhuman_compare}. It can be seen that the performance of the YOLOv7x model trained on the CrowdHuman dataset has significantly improved in crowded classroom scenes.

\begin{table}
    \small
    \centering \renewcommand\arraystretch{1.25} 
    \caption{YOLOv7x training results on CrowdHuman Dataset.}
    \label{yolov7_crowdhuman}
    \begin{tabular}{p{1.8cm}p{1cm}p{1cm}p{1cm}p{1cm}}
        \hline
        class & P (\%) & R (\%) &  mAP@ 50(\%) &  mAP@ 50:95(\%) \\ \hline
        all & 87.7 & 79.6  & 85.6 & 57.2 \\
        head & 88.7 & 79.7 & 85.4 & 55.0 \\
        full body & 88.3 & 81.1 & 86.9 & 58.5\\
        visible body & 86.2 & 78.0 & 84.5 & 58.1 \\
        \hline
    \end{tabular}
\end{table}

\textbf{Student Teacher Identification}

First, we created a dataset with the identities of students and teachers (Student-Teacher Dataset), collecting a total of 785 videos. We sampled a frame every 25 seconds and labeled a total of 2785 video frames, with a total of 4078 labels. Detailed label information is shown in Table \ref{Student_Teacher_Dataset}.

\begin{table}
    \small
    \centering \renewcommand\arraystretch{1.25} 
    \caption{The statistics of Student-Teacher Dataset}
    \label{Student_Teacher_Dataset}
    \begin{tabular}{p{0.6cm}p{2.2cm}p{1.9cm}p{1.9cm}}
        \hline
          & Number of video frames & The number of labels & Student/Teacher labels  \\ \hline
        Train & 2190 & 3200  & 2264/936 \\
        Val & 595 & 878 & 627/251 \\
        \hline
    \end{tabular}
\end{table}

\begin{figure}[htbp]
\centerline{\includegraphics[width=0.48\textwidth]{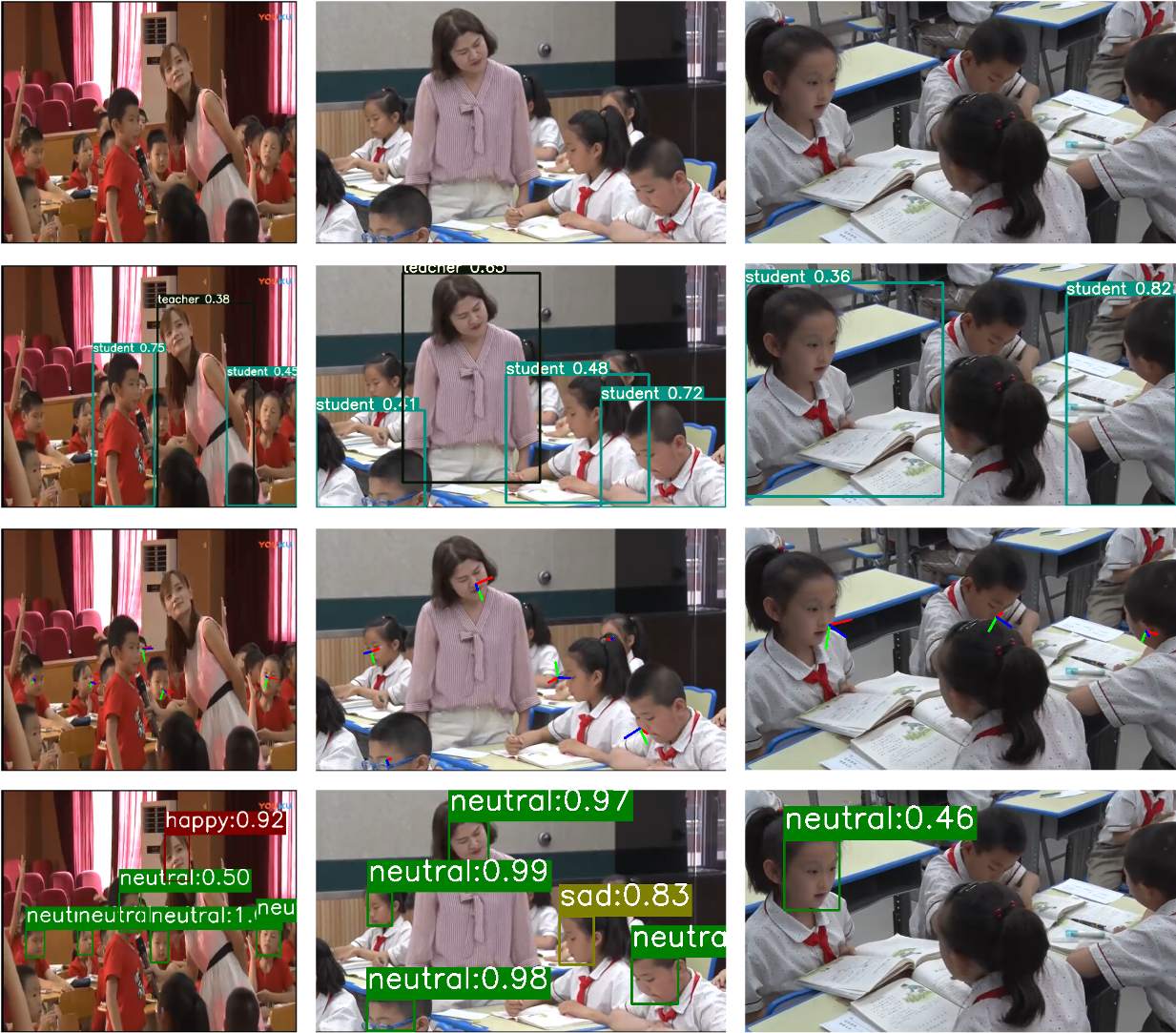}}
\caption{Multi model detection results}
\label{Multi_model_detection_results}
\end{figure}

We trained the Student-Teacher Dataset on YOLOv7, achieving a mAP@50 of 94.5$\%$. Detailed information is shown in Table \ref{YOLOv7_Student_Teacher_Dataset}. The detection results are shown in the second row of Fig.~\ref{Multi_model_detection_results}.

 \begin{table}
    \small
    \centering \renewcommand\arraystretch{1.25} 
    \caption{YOLOv7 training results on Student-Teacher Dataset.}
    \label{YOLOv7_Student_Teacher_Dataset}
    \begin{tabular}{p{1.3cm}p{1.1cm}p{1.1cm}p{1.1cm}p{1.1cm}}
        \hline
          class & P (\%) & R (\%) &  mAP@ 50(\%) &  mAP@ 50:95(\%) \\ \hline
        all & 90.1  & 90.6 & 94.5 & 73.3  \\
        student & 92.4 & 91.6 & 95 & 73.8 \\
        teacher & 87.8 & 89.6 & 93.9 & 72.8  \\
        \hline
    \end{tabular}
\end{table}

The detection results for head pose and facial expression are depicted in the third and fourth rows of Fig.~\ref{Multi_model_detection_results}, respectively.

\section{Conclusion}

In this study, we propose a method to extend image datasets to video datasets. This method not only avoids the high cost of manual annotation, but also enables the field of student behavior detection to enter the realm of video understanding in a faster and more cost-effective way, laying a solid foundation for future research. By combining multiple models, including Deep Sort, YOLOv7, SynergyNet, and Facial Expression, we have obtained crucial data that can contribute to a comprehensive analysis of student behaviors. Finally, we have implemented a process of student concentrated behavior sequence analysis.

Future work includes adding more categories of student behaviors and reducing the data imbalance in the dataset. Additionally, improvements can be made to the SlowFast model from the perspective of loss functions, allowing the network to better learn from fewer samples. Furthermore, the use of state-of-the-art spatiotemporal behavior detection networks, such as MAE networks, can be considered to better integrate object detection and spatiotemporal behavior classification.

\vspace{12pt}


\begin{thebibliography}{00}

\bibitem{Zhu_comprehensive_study} Zhu Y, Li X, Liu C, et al. A comprehensive study of deep video action recognition[J]. arXiv preprint arXiv:2012.06567, 2020.

\bibitem{HUANG_Multi-person_classroom_action} HUANG Y, LIANG M, WANG X, et al. Multi-person classroom action recognition in classroom teaching videos based on deep spatiotemporal residual convolution neural network[J]. Journal of Computer Applications, 2022, 42(3): 736.

\bibitem{He_recognition_student_classroom_behavior} He X, Yang F, Chen Z, et al. The recognition of student classroom behavior based on human skeleton and deep learning[J]. Mod. Educ. Technol, 2020, 30(11): 105-112.

\bibitem{YAN_Student_classroom_behavior} YAN Xing-ya, KUANG Ya-xi, BAI Guang-rui, LI Yue. Student classroom behavior recognition method based on deep learning[J]. Computer Engineering, doi: 10.19678/j.issn.1000-3428.0065369.

\bibitem{Ava} Gu C, Sun C, Ross D A, et al. Ava: A video dataset of spatio-temporally localized atomic visual actions[C]//Proceedings of the IEEE conference on computer vision and pattern recognition. 2018: 6047-6056.

\bibitem{Slowfast} Feichtenhofer C, Fan H, Malik J, et al. Slowfast networks for video recognition[C]//Proceedings of the IEEE/CVF international conference on computer vision. 2019: 6202-6211.

\bibitem{MAE} Feichtenhofer C, Li Y, He K. Masked autoencoders as spatiotemporal learners[J]. Advances in neural information processing systems, 2022, 35: 35946-35958.

\bibitem{UCF101} Soomro K, Zamir A R, Shah M. UCF101: A dataset of 101 human actions classes from videos in the wild[J]. arXiv preprint arXiv:1212.0402, 2012.

\bibitem{kinetics} Carreira J, Zisserman A. Quo vadis, action recognition? a new model and the kinetics dataset[C]//proceedings of the IEEE Conference on Computer Vision and Pattern Recognition. 2017: 6299-6308.

\bibitem{AVA-kinetics} Li A, Thotakuri M, Ross D A, et al. The AVA-kinetics localized human actions video dataset (2020)[J]. arXiv preprint arXiv:2005.00214, 2005.

\bibitem{Multisports} Li Y, Chen L, He R, et al. Multisports: A multi-person video dataset of spatio-temporally localized sports actions[C]//Proceedings of the IEEE/CVF International Conference on Computer Vision. 2021: 13536-13545.

\bibitem{YOLOv5}Jocher, G. (2020). YOLOv5 by Ultralytics (Version 7.0) [Computer software]. https://doi.org/10.5281/zenodo.3908559

\bibitem{YOLOv7} Wang C Y, Bochkovskiy A, Liao H Y M. YOLOv7: Trainable bag-of-freebies sets new state-of-the-art for real-time object detectors[J]. arXiv preprint arXiv:2207.02696, 2022.

\bibitem{YOLOv8} Jocher, G., Chaurasia, A., $\&$ Qiu, J. (2023). YOLO by Ultralytics (Version 8.0.0) [Computer software]. https://github.com/ultralytics/ultralytics

\bibitem{SCB-Dataset1} Fan Y. SCB-dataset: A Dataset for Detecting Student Classroom Behavior[J]. arXiv preprint arXiv:2304.02488, 2023.

\bibitem{SCB-Dataset1-yolov7} Yang F, Wang T, Wang X. Student Classroom Behavior Detection based on YOLOv7-BRA and Multi-Model Fusion[J]. arXiv preprint arXiv:2305.07825, 2023.

\bibitem{SCB-Dataset2}Yang F. Student Classroom Behavior Detection based on Improved YOLOv7[J]. arXiv preprint arXiv:2306.03318, 2023.

\bibitem{SCB-Dataset3} Yang F, Wang T. SCB-Dataset3: A Benchmark for Detecting Student Classroom Behavior[J]. arXiv preprint arXiv:2310.02522, 2023.

\bibitem{Fu_behavior_analysis}Fu R, Wu T, Luo Z, et al. Learning behavior analysis in classroom based on deep learning[C]//2019 Tenth International Conference on Intelligent Control and Information Processing (ICICIP). IEEE, 2019: 206-212.

\bibitem{Zheng_student_behavior}Zheng R, Jiang F, Shen R. Intelligent student behavior analysis system for real classrooms[C]//ICASSP 2020-2020 IEEE International Conference on Acoustics, Speech and Signal Processing (ICASSP). IEEE, 2020: 9244-9248.

\bibitem{Sun_Student_Class_Behavior_Dataset}Sun B, Wu Y, Zhao K, et al. Student Class Behavior Dataset: a video dataset for recognizing, detecting, and captioning students’ behaviors in classroom scenes[J]. Neural Computing and Applications, 2021, 33: 8335-8354.

\bibitem{ACAM}Ulutan O, Rallapalli S, Srivatsa M, et al. Actor conditioned attention maps for video action detection[C]//Proceedings of the IEEE/CVF Winter Conference on Applications of Computer Vision. 2020: 527-536.

\bibitem{MOC}Li Y, Wang Z, Wang L, et al. Actions as moving points[C]//Computer Vision–ECCV 2020: 16th European Conference, Glasgow, UK, August 23–28, 2020, Proceedings, Part XVI 16. Springer International Publishing, 2020: 68-84.

\bibitem{Bsn}Lin T, Zhao X, Su H, et al. Bsn: Boundary sensitive network for temporal action proposal generation[C]//Proceedings of the European conference on computer vision (ECCV). 2018: 3-19.

\bibitem{DBG}Lin C, Li J, Wang Y, et al. Fast learning of temporal action proposal via dense boundary generator[C]//Proceedings of the AAAI conference on artificial intelligence. 2020, 34(07): 11499-11506.

\bibitem{RecNet}Wang B, Ma L, Zhang W, et al. Reconstruction network for video captioning[C]//Proceedings of the IEEE conference on computer vision and pattern recognition. 2018: 7622-7631.

\bibitem{HACA}Wang X, Wang Y F, Wang W Y. Watch, listen, and describe: Globally and locally aligned cross-modal attentions for video captioning[J]. arXiv preprint arXiv:1804.05448, 2018.

\bibitem{Zhou_Classroom_Learning}Zhou J, Ran F, Li G, et al. Classroom Learning Status Assessment Based on Deep Learning[J]. Mathematical Problems in Engineering, 2022, 2022: 1-9.

\bibitem{Li_Student_behavior}Li Y, Qi X, Saudagar A K J, et al. Student behavior recognition for interaction detection in the classroom environment[J]. Image and Vision Computing, 2023: 104726.

\bibitem{Trabelsi_Classroom}Trabelsi Z, Alnajjar F, Parambil M M A, et al. Real-Time Attention Monitoring System for Classroom: A Deep Learning Approach for Student’s Behavior Recognition[J]. Big Data and Cognitive Computing, 2023, 7(1): 48.

\bibitem{Zhou_Stuart}Zhou H, Jiang F, Si J, et al. Stuart: Individualized Classroom Observation of Students with Automatic Behavior Recognition And Tracking[C]//ICASSP 2023-2023 IEEE International Conference on Acoustics, Speech and Signal Processing (ICASSP). IEEE, 2023: 1-5.

\bibitem{BiTNet}Zhao J, Zhu H, Niu L. BiTNet: A lightweight object detection network for real-time classroom behavior recognition with transformer and bi-directional pyramid network[J]. Journal of King Saud University-Computer and Information Sciences, 2023, 35(8): 101670.

\bibitem{CBPH-Net}Zhao J, Zhu H. CBPH-Net: A Small Object Detector for Behavior Recognition in Classroom Scenarios[J]. IEEE Transactions on Instrumentation and Measurement, 2023.

\bibitem{Crowdhuman}
Shao S, Zhao Z, Li B, et al. Crowdhuman: A benchmark for detecting human in a crowd[J]. arXiv preprint arXiv:1805.00123, 2018.

\bibitem{SynergyNet}
Wu C Y, Xu Q, Neumann U. Synergy between 3dmm and 3d landmarks for accurate 3d facial geometry[C]//2021 International Conference on 3D Vision (3DV). IEEE, 2021: 453-463.

\end{thebibliography}
\end{document}